\documentclass[10pt,twocolumn,letterpaper]{article}
\usepackage[pagenumbers]{chordstyle}
\usepackage{times,amsmath,amssymb,amsthm,bm}
\usepackage{graphicx,booktabs,multirow,xcolor,capt-of,colortbl}
\definecolor{linkblue}{rgb}{0.21,0.49,0.74}
\definecolor{rowgray}{gray}{0.94}
\usepackage[pagebackref,breaklinks,colorlinks,allcolors=linkblue]{hyperref}
\usepackage[capitalize]{cleveref}
\newcommand{\R}{\mathbb{R}}
\newcommand{\N}{\mathcal{N}}
\newcommand{\Draft}{\textup{\textsc{Draft}}}
\newcommand{\Full}{\textup{\textsc{Full}}}
\providecommand{\ie}{i.e.,\ }

\newcommand{\ours}{\textup{CHORD}}

\definecolor{tabfirst}{HTML}{F7C5C0}
\definecolor{tabsecond}{HTML}{FDE6C6}
\newcommand{\fst}[1]{\cellcolor{tabfirst}\textbf{#1}}
\newcommand{\snd}[1]{\cellcolor{tabsecond}#1}

\title{\ours{}: Closed-loop Human-Object Rectified-flow Decomposition}
\author{Yi Wang \qquad Yubo Zhao \qquad Chi-Keung Tang\\
The Hong Kong University of Science and Technology\\
{\tt\small ywbu@connect.ust.hk}}

\begin{document}
\raggedbottom
\twocolumn[{%
\renewcommand\twocolumn[1][]{#1}%
\maketitle
\vspace{-6mm}
\begin{center}
\includegraphics[width=\textwidth]{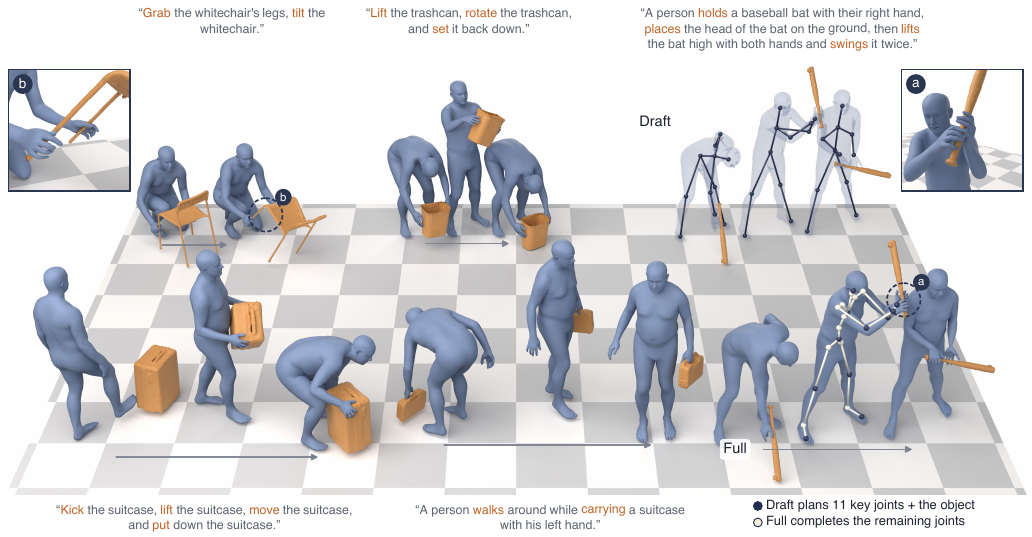}
\captionof{figure}{\ours{} generates whole-body interactions with diverse objects from text; snapshots are ordered in time (gray arrows) and action verbs are in bold. Right column: the two stages for one instruction: the \textbf{Draft} plans 11 key joints with the object, and the \textbf{Full} stage generates the whole body, including fingers, around this plan. Close-ups: both hands (a) around the bat and (b) on the chair legs.}
\label{fig:teaser}
\end{center}
\vspace{2mm}
}]

\begin{abstract}
Text-driven 3D human-object interaction (HOI) generation requires synthesizing whole-body human and object motions that stay coupled through contact. Yet existing generators either entangle the core variables that govern an interaction---object pose, contacting regions, and limb posture---with dozens of subordinate joints, or plan only end-effectors or coarse trajectories. No stage of generation holds the interaction in a form that is both settled and observed, which leaves contact to external guidance, post-hoc refiners, or internal guidance without a surface measurement. We present \ours{}, a two-stage rectified-flow framework that settles the interaction in a compact plan and observes that plan against the object surface while it is being generated. First, \textsc{Draft} uses twelve part tokens per temporal patch to jointly plan the 6-DoF object trajectory, contact states, and the end and middle joints of every kinematic chain. At each flow step, \textsc{Draft} decodes the expected interaction from the closed-form clean estimate, measures it against the object surface, and routes this feedback to the responsible part token with zero extra network evaluations. Second, \textsc{Full} keeps this plan unchanged and infills the remaining body and finger joints. On InterAct, \ours{} reduces FID by 33\% and lifts hand contact F1 from 0.730 to 0.834, while its feedback improves contact on unseen objects and transfers plug-and-play to existing HOI generators.
\end{abstract}
\section{Introduction}
\label{sec:intro}

Generating 3D human-object interactions (HOI) from natural language is a core capability for character animation, virtual reality and embodied AI~\cite{li2024chois,xu2025interact,light2026,dechoi}. Compared with text-to-motion synthesis~\cite{tevet2023mdm,xie2024omnicontrol} or grasping of static objects~\cite{taheri2020grab,taheri2022goal}, text-driven HOI is more demanding: it must jointly synthesize a 6-DoF object trajectory and whole-body human motion, including articulated fingers, that stay coupled through contact throughout the interaction.

Consider the instruction \emph{``lift the bat high with both hands and swing it''} (\cref{fig:teaser}, right). The text leaves three forms of coordination implicit. First, \emph{body parts divide the labor}: the hands grip the handle, the torso rotates into the swing, and the legs stay planted to brace the motion. Second, \emph{body and object move as one}: the bat must track the hands frame by frame rather than drift beside them. Third, \emph{contact persists}: once the hands close around the handle, they must stay on it without hovering, slipping or penetrating.

Recent methods have made notable progress toward these requirements along two axes. The first is how the human and the object are coordinated. Single-stage generators synthesize both together, by coupling a human and an object branch through cross-attention~\cite{peng2023hoidiff}, by learning motion jointly with contact~\cite{diller2024cghoi,xu2025interact}, or by denoising body, hand and object token groups at separate noise levels~\cite{light2026}. Coarse-to-fine pipelines first produce a sparse guide and then synthesize the detailed motion conditioned on it: hand positions predicted from the object motion~\cite{li2023omomo} or coarse human and object trajectories~\cite{dechoi}. The second axis is how contact errors are corrected. Most methods correct them outside the generator. During sampling, guidance steers the sample toward predicted contacts, contact constraints or a learned interaction field~\cite{peng2023hoidiff,diller2024cghoi,li2024chois,kulkarni2024nifty,xu2025interact}, or a separate physics-informed predictor or relation model corrects each denoised estimate~\cite{xu2023interdiff,xue2025rog}. After sampling, the motion is refined by contact optimization or a dedicated refinement network~\cite{grady2021contactopt,zhou2022toch,cha2024text2hoi,liu2024geneoh}. A recent alternative keeps the guidance inside the generator: the body, hands and object are denoised at different paces, so that the parts that are already clean guide those that are still noisy~\cite{light2026}.

Despite these advances, body and object still drift apart during the interaction because no stage of generation holds the interaction in a form that is both settled and observed. It is not settled because the variables that decide it---how the object moves, which body regions touch it, and how each limb is postured---have no dedicated locus. Single-stage generators mix them with every other degree of freedom, while coarse-to-fine pipelines plan only end-effectors or coarse trajectories and leave limb posture open. It is not observed because the generator receives no measurement of where its evolving prediction stands relative to the object surface, which leaves contact to corrections outside the generator that add inference cost, or to internal guidance that lacks such a measurement.

\begin{figure*}[t]
\centering
\includegraphics[width=\textwidth]{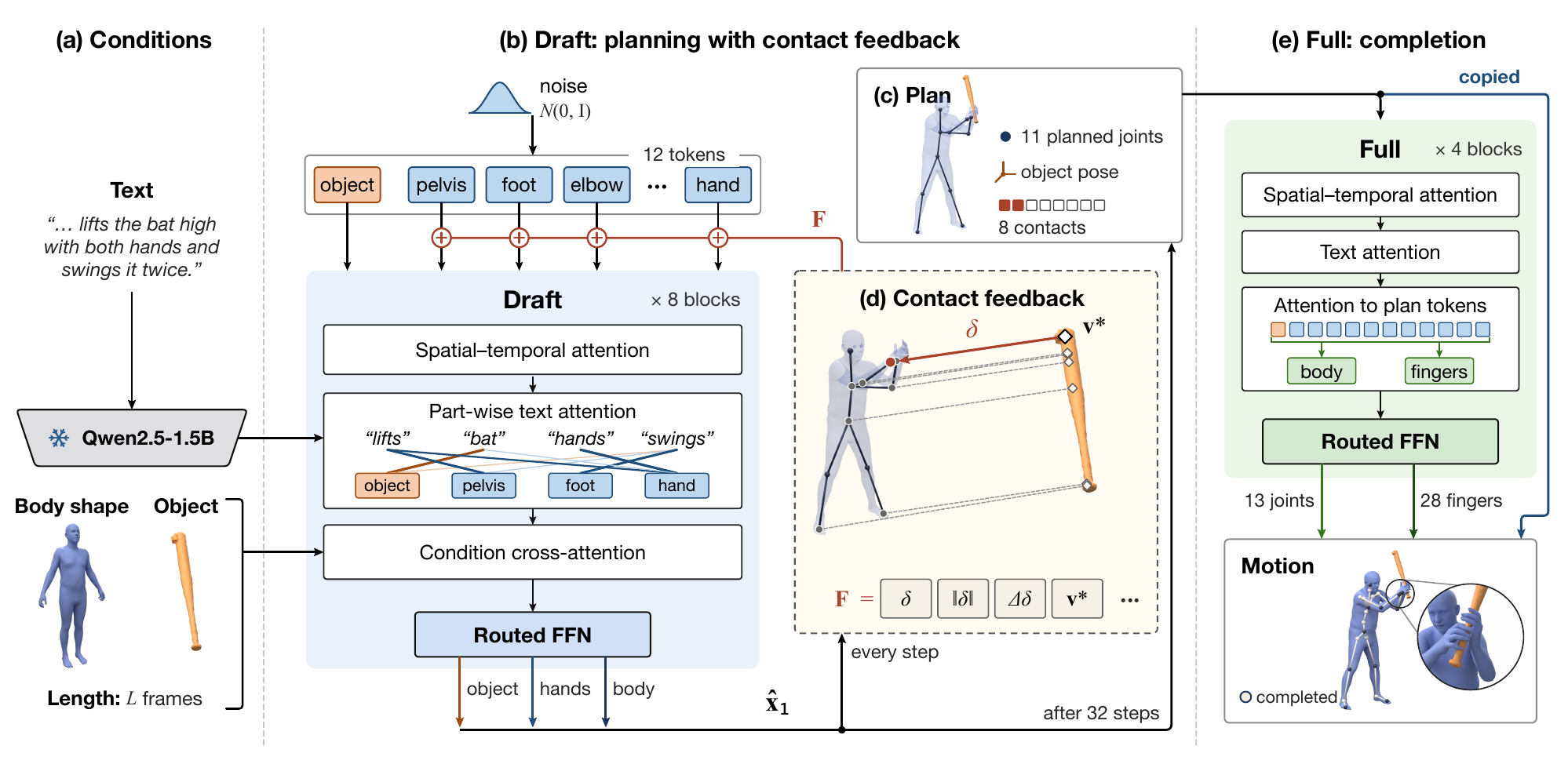}
\vspace{-25pt}
\caption{\textbf{Overview of \ours{}.} (a) Given a text instruction, an object mesh and a body shape, (b) \Draft{} represents each temporal patch with twelve part tokens that jointly generate (c) the plan: the 6-DoF object pose, eleven planned joints---the end and middle joints of every kinematic chain---and binary contact states. (d) At every flow step, the interaction the plan currently expects is decoded from the clean estimate, measured against the object surface, and fed back into the corresponding part token at the next step. (e) \Full{} copies the planned object pose and joints and completes the remaining body and finger joints.}
\label{fig:overview}
\vspace{-2mm}
\end{figure*}

To this end, we present \textbf{\ours{}}, a part-structured rectified-flow framework that decides the interaction in a plan and checks that plan against the object surface while it is being generated (\cref{fig:overview}). To settle the interaction, \ours{} decides it in a dedicated planning stage. Once the object pose, the pelvis, and the end and middle joints of every kinematic chain are fixed, the remaining body joints are tightly constrained by skeletal geometry. Our first stage, \Draft{}, therefore represents each temporal patch with twelve part tokens, one for the object and one per planned joint, that jointly generate the object trajectory, these joints and their contact states. 
Our second stage, \Full{}, takes this plan as fixed and infills the remaining body joints and the fingers.

To observe it, \ours{} closes the contact loop inside the learned velocity field of \Draft{}. At every flow step, the closed-form clean estimate of the rectified flow~\cite{liu2023rectified,lipman2023flow} exposes the interaction the plan currently expects to produce. We measure this expected interaction against the object surface and return the measurement to the part token responsible for it, so the plan is revised while it is being formed, without extra network evaluations, as the measurement is computed from the clean estimate the velocity field already produces. By deciding the interaction in one plan and checking it as it forms, \ours{} keeps body and object coupled throughout the interaction.

On the InterAct benchmark~\cite{xu2025interact}, \ours{} outperforms prior text-to-HOI generators in motion realism and contact accuracy. The feedback keeps improving contact on objects unseen in training. We further demonstrate its versatility as a plug-and-play module, which improves contact accuracy and reduces floating objects when integrated into existing HOI generators~\cite{light2026,xu2025interact}.

Our contributions are summarized as follows:
\begin{itemize}
    \item We propose \ours{}, a two-stage rectified-flow framework that decides the interaction in a part-structured plan and then completes the remaining body and finger joints.
    \item We introduce in-generation contact feedback, which measures the expected interaction against the object surface at every flow step and corrects contact without gradient guidance or extra network evaluations.
    \item Extensive experiments show that \ours{} achieves state-of-the-art motion realism and contact accuracy, and that its feedback transfers to unseen objects and to existing generators.
\end{itemize}
\section{Related Work}
\label{sec:related}

\paragraph{Human-object interaction generation.}
Early HOI synthesis addressed reaching and grasping static objects~\cite{taheri2020grab,taheri2022goal,wu2022saga,jiang2021grasptta} or forecasting short interactions from past motion~\cite{xu2023interdiff}. Full-body manipulation datasets~\cite{bhatnagar2022behave,huang2022intercap,li2023omomo,xu2025interact} shifted attention to generating human and object motion jointly from text, alongside zero-shot approaches that compose pretrained priors~\cite{xu2024interdreamer,cao2025avatargo,lou2025zeroshot,li2026hoipage,dai2026anchorhoi} and physics-based approaches that realize interactions in simulation~\cite{wang2023physhoi,xu2025intermimic,xu2026interprior,lyric2026}. Kinematic methods further cover intent conditioning, fine-grained semantics, hand-object or multi-body interaction, static human-object configurations, unseen objects and autoregressive or masked generation~\cite{ghosh2023imos,yang2024fhoi,christen2024diffh2o,he2025syncdiff,petrov2025tridi,zhang2024oodhoi,li2025genhoi,geng2025ardhoi,ji2026hoimask}. Among text-driven generators, HOI-Diff~\cite{peng2023hoidiff}, CG-HOI~\cite{diller2024cghoi} and Text2HOI~\cite{cha2024text2hoi} couple human and object generation with learned contact or affordance predictors, and LIGHT~\cite{light2026} denoises the body, hands and object asynchronously.

\paragraph{Hierarchical and part-aware generation.}
To coordinate body and object motion, hierarchical methods split interaction synthesis into stages. OMOMO~\cite{li2023omomo} predicts hand trajectories from a given object motion before generating the full body, CHOIS~\cite{li2024chois} conditions on sparse object waypoints and DecHOI~\cite{dechoi} generates coarse human and object trajectories prior to detailed motion. However, the intermediate representation is either supplied at test time or limited to trajectories and end-effectors that leave limb posture unconstrained. Part-aware conditioning has been explored in object-free motion generation, through part-level descriptions~\cite{finemogen,frankenmotion}, prompt decomposition~\cite{lgtm}, part embeddings derived from a global sentence vector~\cite{party} or coordinated part-wise generators~\cite{zou2024parco}, and in HOI through learnable chain-level queries, shared across the frames of a sequence, that gather the text and object geometry~\cite{zeng2025chainhoi}. In contrast, the plan of \ours{} is generated from text alone, covers the object trajectory, contact states and the end and middle joints of every kinematic chain, and is copied unchanged into the final motion. Each of its part tokens queries the words of the instruction at every temporal patch, so each part attends to the words that concern it as the motion unfolds.

\paragraph{Contact correction.}
Existing pipelines typically correct contact outside the generator. Sampling is steered through classifier or training-free guidance~\cite{dhariwal2021diffusion,chung2023dps,yu2023freedom,bansal2024universal,he2024mpgd,karunratanakul2023gmd,rempe2023trace} with affordance or contact objectives~\cite{peng2023hoidiff,diller2024cghoi,xu2025interact}, neural interaction fields~\cite{kulkarni2024nifty} or physics-informed or relation-based corrections during denoising~\cite{yuan2023physdiff,xu2023interdiff,xue2025rog}. Others apply post-hoc contact optimization~\cite{grady2021contactopt,zhou2022toch}, optimization of the diffusion noise~\cite{karunratanakul2024dno,ron2025hoidini} or a separate refinement network~\cite{liu2024geneoh}. These corrections add gradient steps, optimization or a second network at inference. A recent alternative keeps the guidance inside the generator through asynchronous denoising~\cite{light2026}, which removes the external classifier but does not measure the prediction against the object surface. In contrast, \ours{} measures the expected interaction against the object surface at every flow step and feeds the measurement back into the responsible part token, so the correction becomes part of the velocity field.

\section{Method}
\label{sec:method}

Given only a text instruction $d$, an object mesh with scale $s_{\text{obj}}$ and canonical surface points $\mathcal{V}_{\text{obj}}$, a body shape $\beta$ and a sequence length $L$, \ours{} generates the object trajectory $\mathbf{o}_{1:L}$ (translation and rotation) together with the whole-body motion $\mathbf{J}_{1:L} \in \R^{L \times 52 \times 3}$ ($22$ body and $30$ hand joints). As shown in \cref{fig:overview}, the first stage, \Draft{}, generates a part-structured interaction plan (\cref{subsec:plan}) and checks it against the object at every sampling step (\cref{subsec:feedback}). The second stage, \Full{}, completes the remaining degrees of freedom and leaves the plan unchanged (\cref{subsec:plan}).

\subsection{Preliminaries: Rectified Flow}
\label{subsec:prelim}
Both stages are formulated as rectified-flow models~\cite{liu2023rectified,lipman2023flow}. Given a normalized target sample $\mathbf{x}_1$ and Gaussian noise $\mathbf{x}_0 \sim \N(\mathbf{0}, \mathbf{I})$, the linear interpolant at flow time $s \in [0,1]$ is $\mathbf{x}_s = (1-s)\,\mathbf{x}_0 + s\,\mathbf{x}_1$, and \Draft{} predicts the velocity field $\mathbf{v}_\theta(\mathbf{x}_s, s, \mathcal{C}) \approx \mathbf{x}_1 - \mathbf{x}_0$ under condition $\mathcal{C}$. At any flow step, the corresponding clean-sample estimate is available in closed form:
\begin{equation}
\hat{\mathbf{x}}_1 = \mathbf{x}_s + (1 - s)\,\mathbf{v}_\theta(\mathbf{x}_s, s, \mathcal{C}).
\label{eq:clean}
\end{equation}
This estimate approximates the mean of the final sample given $\mathbf{x}_s$, \ie the interaction the generator currently expects to produce, and it is available at every flow step without an extra network evaluation (\cref{subsec:feedback}). Both stages regress the velocity of the interpolant,
\begin{equation}
\mathcal{L}_{\text{FM}} = \mathbb{E}_{s,\,\mathbf{x}_0,\,\mathbf{x}_1}\,\big\|\mathbf{v}_\theta(\mathbf{x}_s, s, \mathcal{C}) - (\mathbf{x}_1 - \mathbf{x}_0)\big\|^2,
\label{eq:fm}
\end{equation}
and are sampled by Euler integration from $s=0$ to $1$. In \Draft{}, the velocity field additionally receives the feedback $\mathbf{F}$ of \cref{subsec:feedback}, $\mathbf{v}_\theta(\mathbf{x}_s, s, \mathcal{C}, \mathbf{F})$.

\subsection{Part-Structured Planning and Completion}
\label{subsec:plan}

\paragraph{Interaction plan.}
The plan should fix what defines the interaction and leave to \Full{} what skeletal geometry already determines. It therefore holds the object pose, which locates the interaction surface, the pelvis with its root orientation, which places the body, and the end joint together with the middle joint of five kinematic chains: hands and elbows, feet and knees, head and chest. The end joints set where the body reaches and touches; the middle joints fix the limb posture that the end joints leave open and that hand-first plans~\cite{li2023omomo} leave undetermined. Every remaining body joint then moves rigidly with a planned joint or lies on a short bone chain between two of them. The plan thus constrains the body pose up to the fingers, and \Full{} places the remaining joints within these constraints and generates the fingers. At each frame $t$, \Draft{} predicts a 48-D kinematic state alongside binary contact indicators for eight body regions:
\begin{equation}
\begin{aligned}
\mathbf{x}^{\Draft{}}(t) &= \big[\,\mathbf{o}(t),\; \mathbf{p}_{\text{world}}(t),\; \mathbf{r}(t),\; \tilde{\mathbf{p}}_{\text{root}}(t)\,\big] \in \R^{48},\\
\mathbf{c}(t) &\in \{0,1\}^{8}.
\end{aligned}
\label{eq:plan}
\end{equation}
Here $\mathbf{o}\in\R^9$ is the object translation and 6D rotation, $\mathbf{p}_{\text{world}}\in\R^{15}$ the world positions of the pelvis, hands and feet, where object and ground contact occur, $\mathbf{r}\in\R^6$ the root orientation, and $\tilde{\mathbf{p}}_{\text{root}}\in\R^{18}$ the positions of the chest, knees, elbows and head in the pelvis-centered root frame, which encodes posture independently of global position and heading. The eight contact regions---hands, feet, pelvis, back and forearms---cover the body surfaces that contact objects.

\paragraph{Part tokens.}
\Draft{} (\cref{fig:overview}b) groups every $K=4$ frames into a temporal patch and represents each patch with $P=12$ part tokens: one for the object and one per planned joint, with the root orientation in the pelvis token. Each token embeds only its own channels from \cref{eq:plan}, so that each part reads the instruction and receives feedback separately (\cref{eq:query}, \cref{subsec:feedback}). Every transformer block~\cite{vaswani2017attention} applies spatial self-attention across the twelve part tokens of a patch, temporal self-attention along each part's trajectory, and cross-attention to the condition tokens (a basis point set~\cite{prokudin2019bps} encoding of the object geometry, the body shape and the sequence length), modulated by adaptive layer normalization~\cite{peebles2023dit} on the flow time and these conditions. Attention projections are shared across tokens, whereas the feed-forward network is routed by token type (object, hands, body). Each token decodes its $K$ frames through a linear head, and each region's contact logits are predicted by a classification head on its part token, outside the flow state.

\paragraph{Part-wise attention to the instruction.}
The instruction is encoded once by a frozen language model into word features $\mathbf{W} \in \R^{L_w \times d_{\text{lm}}}$ and a part anchor $\mathbf{a}_m$ for each of the twelve tokens, read from that part's name appended to the prompt. In every block, the token of part $m$ at patch $t_p$ forms its own query over the words:
\begin{equation}
\mathbf{q}_{t_p,m} = \mathbf{W}_q\big(\bar{\mathbf{h}}_{t_p,m} + \psi_a(\mathbf{a}_m) + \psi_\tau(t_p)\big),
\label{eq:query}
\end{equation}
where $\bar{\mathbf{h}}_{t_p,m}$ is its modulated hidden state, $\mathbf{W}_q$ is the query projection of the attention, and $\psi_a$ and $\psi_\tau$ are linear embeddings of the part anchor and the patch position. A fixed per-head distance bias between the relative position of patch $t_p$ within the sequence and that of each word within the caption favors temporally corresponding words. Each part thus reads the instruction with its own distribution over words rather than one pooled representation shared by the whole body.

\paragraph{Completion.}
What the plan leaves open does not decide the interaction: the remaining body joints are constrained by the planned ones through skeletal geometry, and the fingers realize contact around hand positions that the plan has already set. \Full{} completes the motion by infilling (\cref{fig:overview}e). It keeps the object trajectory and the eleven planned joints from \Draft{} unchanged and generates only the remaining 13 body joints and 28 finger joints (the two planned hand joints are among the 30 hand joints), with finger positions expressed as offsets from their hand joint. Each frame is represented by a body token and one finger token for both hands. Every block applies attention between the two tokens, temporal attention, cross-attention to the instruction words and to twelve plan tokens that embed the plan of the same frame part by part, and a token-type-routed feed-forward network, with adaptive layer normalization on the flow time, object geometry, body shape and sequence length. The token inputs combine the planned state with the surface relations and contact probabilities predicted by \Draft{}; the finger token additionally receives the object surface points nearest to each planned hand.

\subsection{In-Generation Contact Feedback}
\label{subsec:feedback}

Because \Full{} leaves the planned object trajectory and joints unchanged, whether the body reaches the object is decided in \Draft{}. A generator that encodes object geometry only as a static condition, however, has no explicit measurement of whether the interaction it currently expects to produce reaches the surface. \ours{} instead checks the expected interaction while it is being decided: at every flow step, the interaction the plan currently expects is measured against the object surface, and the measurement is returned to the part token that will revise it (\cref{fig:overview}d).

\paragraph{Expected interaction.}
The clean estimate (\cref{eq:clean}) is decoded into the expected positions $\hat{\mathbf{p}}_m(t)$ of the eight contact regions and the expected object pose $(\hat{\mathbf{R}}(t), \hat{\mathbf{o}}_{\text{trans}}(t))$, and each region is expressed in the expected object frame:
\begin{equation}
\hat{\mathbf{q}}_m(t) = \hat{\mathbf{R}}(t)^\top\big(\hat{\mathbf{p}}_m(t) - \hat{\mathbf{o}}_{\text{trans}}(t)\big).
\label{eq:expected}
\end{equation}
Every contact region is located by planned joints: the hands and feet directly, the pelvis and back rigidly from the pelvis and root orientation, and each forearm at a fixed point between its elbow and hand. The expectation is therefore observable without completing the body.

\paragraph{Surface measurement.}
Let $\mathbf{v}^*_m(t) \in \mathcal{V}_{\text{obj}}$ denote the canonical surface point nearest to $\hat{\mathbf{q}}_m(t)$, and let $\boldsymbol{\delta}_m(t) = \hat{\mathbf{q}}_m(t) - \mathbf{v}^*_m(t)$ be the surface offset. The measurement is
\begin{equation}
\begin{aligned}
\mathbf{f}^{\text{geo}}_m(t) = \big[&\boldsymbol{\delta}_m,\; \|\boldsymbol{\delta}_m\|,\; \boldsymbol{\delta}_m(t) - \boldsymbol{\delta}_m(t-1),\\
&\mathbf{v}^*_m / s_{\text{obj}},\; 1\big] \in \R^{11}
\end{aligned}
\label{eq:feature}
\end{equation}
Each entry tells the part token something it needs to correct its contact. The offset $\boldsymbol{\delta}_m$ and its norm give the direction and size of the gap between the region and the surface, which a hovering region has to close. The frame-to-frame difference is zero for a region that moves rigidly with the object, since all quantities are expressed in the object frame, so a nonzero value exposes a region that drifts or slips on the object. The scale-normalized surface point $\mathbf{v}^*_m / s_{\text{obj}}$ tells where on the object the region currently is, independently of the object's translation and rotation. The final entry marks the measurement as available: when no feedback is given, at the first sampling step and for the training samples without feedback, the whole feature is zero.

\paragraph{Routing.}
The feedback $\mathbf{F}(\hat{\mathbf{x}}_1)$ stacks each region's measurement and a contact-affordance feature over the $K$ frames of a patch and injects it through zero-initialized linear pathways at the token embedding and before every block. Hand, foot and pelvis measurements enter their own tokens, the back measurement enters the chest token and the forearm measurements enter the elbow tokens. The measurements condition the subsequent updates of the velocity field together with the instruction.

\subsection{Training and Inference}
\label{subsec:training}

\paragraph{Objective.}
\Draft{} is trained with
\begin{equation}
\begin{aligned}
\mathcal{L}_{\Draft{}} = \frac{1}{|\mathcal{G}|}\sum_{g \in \mathcal{G}} \mathcal{L}^{g}_{\text{FM}} &+ \lambda_{c}\,\mathcal{L}_{\text{con}} + \lambda_{r}\,\mathcal{L}_{\text{rel}} \\
&+ \lambda_{l}\,\mathcal{L}_{\text{limb}} + \lambda_{f}\,\mathcal{L}_{\text{foot}} + \lambda_{o}\,\mathcal{L}_{\text{obj}},
\end{aligned}
\label{eq:loss}
\end{equation}
where $\mathcal{L}^{g}_{\text{FM}}$ is the velocity loss of \cref{eq:fm} restricted to channel group $g$ of the plan (object translation, object rotation, pelvis, hands, feet, root orientation and root-frame joints) and $\mathcal{L}_{\text{con}}$ is a binary cross-entropy on the contact logits. The remaining terms are evaluated on the clean estimate $\hat{\mathbf{x}}_1$. $\mathcal{L}_{\text{rel}}$ anchors each contacting region to its ground-truth position in the object frame (\cref{eq:expected}) on ground-truth contact frames. $\mathcal{L}_{\text{limb}}$ preserves limb lengths, $\mathcal{L}_{\text{foot}}$ penalizes foot motion on stance frames and $\mathcal{L}_{\text{obj}}$ penalizes object motion on static-object frames. \Full{} is trained separately on noise-perturbed ground-truth plans with $\mathcal{L}_{\text{FM}}$ on its own channels plus a contact and a foot-skating term.

\paragraph{Two-pass training.}
To learn from feedback computed on its own predictions, \Draft{} is trained in two passes. On $90\%$ of training samples, a stop-gradient first pass is evaluated without feedback on the same noise and target at an earlier, noisier flow time $s_0 = \max(0, s - \Delta s)$ with a fixed shift $\Delta s$. Its feedback $\mathbf{F}(\hat{\mathbf{x}}_1(s_0))$ then conditions the second pass at time $s$, which receives the gradient. Using an earlier flow time mirrors sampling, where the feedback comes from the previous Euler step. The remaining $10\%$ train without feedback, matching the first sampling step.

\paragraph{Inference.}
\Draft{} integrates the flow over $N$ Euler steps $s_k = k/N$ with classifier-free guidance~\cite{ho2022cfg} on the instruction. Step $k$ reuses the feedback of the previous step's expected interaction,
\begin{equation}
\mathbf{x}_{s_{k+1}} = \mathbf{x}_{s_k} + \tfrac{1}{N}\,\tilde{\mathbf{v}}_\theta\big(\mathbf{x}_{s_k}, s_k, \mathcal{C}, \mathbf{F}(\hat{\mathbf{x}}_1^{(k-1)})\big),
\label{eq:sample}
\end{equation}
where $\tilde{\mathbf{v}}_\theta$ is the guided velocity and $\mathbf{F}(\hat{\mathbf{x}}_1^{(-1)}) = \mathbf{0}$ for the first step. The clean estimate $\hat{\mathbf{x}}_1^{(k)}$ is formed from the guided velocity of step $k$ (\cref{eq:clean}), so the feedback reuses that step's evaluations. \Full{} then completes the motion with its own Euler integration.

\section{Experiments}
\label{sec:experiments}

\subsection{Experimental Setup}
\label{subsec:exp_setup}

\paragraph{Dataset.}
We evaluate on InterAct~\cite{xu2025interact}, a benchmark of whole-body human-object interactions with text descriptions and object meshes. We follow its official training and test splits. 

\paragraph{Metrics and protocol.}
We measure alignment and realism via retrieval precision (R-Precision, top-1/2/3), FID, multimodal distance (MM Dist) and diversity~\cite{guo2022humanml3d}; physical plausibility via the foot-skating ratio (FSR, the share of frames with foot sliding) and object penetration (Pene, the mean depth of body joints inside the object); and interaction accuracy via the contact ratio (Contact, the share of frames in contact) and frame-wise hand contact precision, recall and F1 against the ground truth. All methods are scored with the pretrained evaluator of LIGHT~\cite{light2026}. In ablations, $\text{F1}_H$ denotes hand contact F1 (\ie $C_{F1}$) and R@3$_{256}$ top-3 R-Precision at batch size 256.

\paragraph{Baselines.}
We compare with HOI-Diff~\cite{peng2023hoidiff}, CHOIS~\cite{li2024chois}, Text2HOI~\cite{cha2024text2hoi}, InterAct~\cite{xu2025interact}, ROG~\cite{xue2025rog} and LIGHT~\cite{light2026}. All are trained on InterAct with text, object geometry and body shape as the only inputs, CHOIS without waypoints, and scored with the same evaluator.

\paragraph{Implementation details.}
Object surfaces are sampled with $4096$ points, and clips have at most $300$ frames at $30$ fps. The instruction is encoded by a frozen \texttt{Qwen2.5-1.5B}~\cite{qwen25}, and \Draft{} uses classifier-free guidance scale $0.5$. Our model and every ablation variant are trained with the same recipe for 170k updates, and their results are means over ten sampling seeds. Sampling a clip takes $N=32$ \Draft{} and 8 \Full{} Euler steps, 72 network evaluations under classifier-free guidance, without gradient guidance or an auxiliary network.

\subsection{Comparison with Prior Methods}
\label{subsec:exp_main}

\paragraph{Quantitative comparison.}
As shown in \cref{tab:main}, \ours{} achieves the lowest FID, MM Dist and penetration, the best top-2 and top-3 R-Precision and the diversity closest to the ground truth. It also has the highest contact precision, recall and F1. Both designs contribute to these gains, as the ablations in \cref{subsec:exp_plan,subsec:exp_feedback} show. On each of the six subsets, \ours{} also has the best contact F1, R-Precision and MM Dist against InterAct and LIGHT.

\begin{figure*}[t]
\centering
\includegraphics[width=\textwidth]{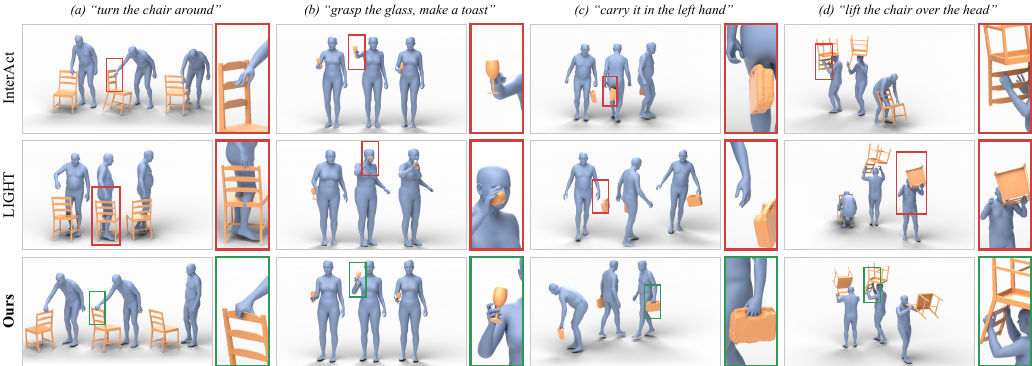}
\vspace{-2mm}
\caption{\textbf{Qualitative comparison on InterAct.} Each panel shows three frames of one sample, left to right in time. Close-ups mark baseline errors (red) and our contacts (green).}
\label{fig:qual}
\vspace{-2mm}
\end{figure*}

\paragraph{Qualitative comparison.}
\Cref{fig:qual} compares representative sequences with the released InterAct~\cite{xu2025interact} and LIGHT models. The baselines move the object without touching it, leave it floating between the hands or let the body pass through it. In our sequences, the hands stay on the object while it moves.

\begin{table*}[t]
\centering
\caption{\textbf{Comparison on InterAct.} R-Precision uses batch size 256; Pene in mm; ground-truth diversity $7.810$ and contact $0.208$. Best and second best are shaded red and orange.}
\label{tab:main}
\vspace{-1.5mm}
\resizebox{\textwidth}{!}{%
\begin{tabular}{l|ccc|c|c|c|c|c|c|ccc}
\toprule
\multirow{2}{*}{\textbf{Method}} & \multicolumn{3}{c|}{\textbf{R-Precision}$\uparrow$} & \multirow{2}{*}{\textbf{FID}$\downarrow$} & \multirow{2}{*}{\textbf{MM Dist}$\downarrow$} & \multirow{2}{*}{\textbf{Diversity}$\rightarrow$} & \multirow{2}{*}{\textbf{FSR}$\downarrow$} & \multirow{2}{*}{\textbf{Pene}$\downarrow$} & \multirow{2}{*}{\textbf{Contact}$\rightarrow$} & \multicolumn{3}{c}{\textbf{Interaction}$\uparrow$} \\
& Top 1 & Top 2 & Top 3 & & & & & & & $C_{\text{prec}}$ & $C_{\text{rec}}$ & $C_{F1}$ \\
\midrule
HOI-Diff~\cite{peng2023hoidiff} & 0.415 & 0.622 & 0.735 & 0.685 & 3.015 & 7.622 & 0.073 & 1.051 & 0.085 & 0.725 & 0.451 & 0.507 \\
CHOIS~\cite{li2024chois} & \fst{0.434} & \snd{0.646} & \snd{0.763} & 0.583 & 2.772 & 7.707 & 0.116 & 1.213 & 0.117 & 0.715 & 0.522 & 0.546 \\
Text2HOI~\cite{cha2024text2hoi} & 0.429 & 0.640 & 0.743 & 0.315 & 2.637 & 7.625 & 0.065 & 1.042 & 0.105 & 0.709 & 0.483 & 0.527 \\
InterAct~\cite{xu2025interact} & 0.358 & 0.575 & 0.692 & 0.322 & 2.675 & \snd{7.765} & 0.071 & 1.463 & 0.160 & 0.758 & \snd{0.723} & 0.713 \\
ROG~\cite{xue2025rog} & 0.412 & 0.625 & 0.739 & 0.323 & 2.554 & 7.758 & \fst{0.059} & 1.057 & \snd{0.170} & 0.783 & 0.645 & 0.663 \\
LIGHT~\cite{light2026} & 0.346 & 0.555 & 0.690 & \snd{0.170} & \snd{2.427} & 7.673 & 0.069 & \snd{0.953} & 0.141 & \snd{0.830} & 0.698 & \snd{0.730} \\
\midrule
\textbf{\ours{} (ours)} & \snd{0.431} & \fst{0.650} & \fst{0.779} & \fst{0.114} & \fst{2.154} & \fst{7.798} & \snd{0.062} & \fst{0.941} & \fst{0.190} & \fst{0.855} & \fst{0.846} & \fst{0.834} \\
\bottomrule
\end{tabular}}
\vspace{-2mm}
\end{table*}

\paragraph{Unseen objects.}
To evaluate generalization to unseen shapes, we follow the protocol of DecHOI~\cite{dechoi} and replace the object in OMOMO~\cite{li2023omomo} test clips with 17 meshes from 3D-FUTURE~\cite{fu20213dfuture} of the same categories, none of which is seen in training, while keeping the captions (\cref{tab:unseen}). \ours{} follows the instruction more closely than the released LIGHT model, makes more hand contact, leaves the object floating in fewer frames, breaks contact less often and keeps the hands closer to the moving object. Disabling the feedback at inference lowers contact and raises floating and contact breaks on these shapes.

\begin{table}[t]
\setlength{\abovecaptionskip}{2pt}\setlength{\belowcaptionskip}{0pt}
\centering
\caption{\textbf{Unseen 3D-FUTURE objects.} R-Precision in batches of 32; Contact: frames with a hand joint within 5\,cm of the object surface; Floating: frames in which the object moves while no hand is in contact; Breaks: contact interruptions per hand; $D_{\text{hand}}$: hand-to-surface distance (mm) while the object moves. $^\dagger$Released model.}
\label{tab:unseen}
\vspace{0.5mm}
\setlength{\tabcolsep}{3pt}%
\resizebox{\columnwidth}{!}{%
\begin{tabular}{l|cc|c|c|c|c|c}
\toprule
\textbf{Method} & \textbf{R@1}$\uparrow$ & \textbf{R@3}$\uparrow$ & \textbf{MM Dist}$\downarrow$ & \textbf{Contact (\%)}$\uparrow$ & \textbf{Floating (\%)}$\downarrow$ & \textbf{Breaks}$\downarrow$ & $\boldsymbol{D}_{\textbf{hand}}\downarrow$ \\
\midrule
LIGHT$^\dagger$ & .578 & .815 & 4.01 & 53.4 & 24.9 & \snd{0.99} & 89.0 \\
Ours w/o feedback & \snd{.706} & \snd{.887} & \fst{3.53} & \snd{60.6} & \snd{16.9} & 1.03 & \snd{73.4} \\
\ours{} & \fst{.709} & \fst{.888} & \fst{3.53} & \fst{66.1} & \fst{12.1} & \fst{0.77} & \fst{64.9} \\
\bottomrule
\end{tabular}}
\vspace{-3.5mm}
\end{table}

\subsection{Part-Structured Plan}
\label{subsec:exp_plan}

\paragraph{Impact of part tokens.}
To test whether the plan needs one token per part, we replace the twelve part tokens with a single whole-body token per patch, row~(a), or with four coarse tokens for the object, body, hands and feet, row~(b) (\cref{tab:abl_plan}). Both variants are worse in FID, retrieval, MM Dist, penetration and hand contact, the single token most. Row~(c) shares one FFN across all token types to test the type-routed FFN, which also worsens these metrics, penetration in particular.

\begin{table}[t]
\centering
\caption{\textbf{Ablation of the part-structured plan.} (a) One whole-body token per patch instead of twelve part tokens; (b) four tokens (object, body, hands, feet); (c) one FFN shared by all token types; (d) \Full{} regenerates the planned joints and object pose instead of copying them; (e) the \Draft{} backbone generates the full state in one stage, without \Full{}. Last row: \Full{} completes ground-truth plans. Pene in mm; ground-truth diversity $7.810$.}
\label{tab:abl_plan}
\vspace{-1.5mm}
\setlength{\tabcolsep}{3pt}%
\resizebox{\columnwidth}{!}{%
\begin{tabular}{l|ccccc|ccc}
\toprule
\textbf{Variant} & \textbf{FID}$\downarrow$ & \textbf{R@3}$_{256}\uparrow$ & \textbf{MM Dist}$\downarrow$ & \textbf{Div.}$\rightarrow$ & \textbf{Pene}$\downarrow$ & $\boldsymbol{C}_{\textbf{prec}}\uparrow$ & $\boldsymbol{C}_{\textbf{rec}}\uparrow$ & $\textbf{F1}_H\uparrow$ \\
\midrule
Ours (12 part tokens) & \fst{0.114} & \fst{0.779} & \fst{2.154} & 7.798 & \fst{0.941} & \fst{0.855} & \fst{0.846} & \fst{0.834} \\
(a) 1 token & 0.149 & 0.755 & 2.282 & 7.822 & 1.16 & 0.810 & 0.780 & 0.778 \\
(b) 4 tokens & 0.136 & 0.769 & 2.204 & 7.819 & 1.08 & 0.832 & 0.811 & 0.801 \\
(c) w/o routed FFN & \snd{0.129} & 0.774 & 2.188 & \snd{7.809} & 1.15 & \snd{0.851} & \snd{0.829} & \snd{0.818} \\
(d) plan re-predicted & 0.133 & \snd{0.775} & \snd{2.187} & \fst{7.810} & \snd{1.01} & 0.841 & 0.812 & 0.805 \\
(e) one stage & 0.205 & 0.761 & 2.297 & \snd{7.811} & 1.19 & 0.813 & 0.754 & 0.760 \\
\midrule
\rowcolor{rowgray}
Ground-truth plan & 0.000 & 0.848 & 1.493 & 7.810 & 0.72 & 0.984 & 0.983 & 0.982 \\
\bottomrule
\end{tabular}}
\vspace{-2mm}
\end{table}

\paragraph{Impact of two-stage planning.}
To test whether a separate planning stage is needed, row~(e) lets the \Draft{} backbone generate the full state in one stage, with the same tokens, feedback and sampler. This variant is worse in FID, retrieval, MM Dist, penetration and every contact metric. Deciding the interaction in a compact plan and completing the rest afterwards is itself a source of the gain.

\paragraph{Impact of preserving the plan.}
To test whether the second stage should be free to change the plan, row~(d) lets \Full{} regenerate the object pose and planned joints. Regeneration is worse in FID, retrieval, MM Dist, penetration and contact. To test whether the plan alone settles the interaction, we also run \Full{} on ground-truth plans. It reproduces the ground-truth motion distribution and recovers nearly all ground-truth hand contact. Under generated plans, most of the remaining contact error lies in where the plan places the hands, which is where the feedback of \cref{subsec:exp_feedback} operates.

\subsection{In-Generation Contact Feedback}
\label{subsec:exp_feedback}

\paragraph{Impact of contact feedback.}
To measure the effect of the feedback, we train and sample models without it, with self-conditioning~\cite{chen2023analogbits} on each token's own clean estimate in its place, with only one of its two injection pathways, and with the same measurements broadcast to all twelve tokens instead of routed (\cref{tab:abl_feedback}). Feedback improves every metric over the model without it, and the two pathways together work better than either alone. Self-conditioning improves FID but only changes retrieval and contact a little. The contact gain comes from the surface measurement. The unrouted variant performs worse than routed feedback across all metrics, including retrieval, hand contact, and FID. Routing each measurement to the token of its own part gives the lowest FID and MM Dist and  highest contact precision, recall and F1.

\begin{table}[t]
\setlength{\abovecaptionskip}{2pt}\setlength{\belowcaptionskip}{0pt}
\centering
\caption{\textbf{Ablation of contact feedback.} Feedback is removed (w/o feedback), injected only into the token embedding (token only) or only before every block (block only), broadcast to all tokens (unrouted), or replaced by each token's own clean estimate (self-conditioning).}
\label{tab:abl_feedback}
\vspace{0.5mm}
\setlength{\tabcolsep}{3pt}%
\resizebox{\columnwidth}{!}{%
\begin{tabular}{l|ccc|ccc}
\toprule
\textbf{Variant} & \textbf{FID}$\downarrow$ & \textbf{MM Dist}$\downarrow$ & \textbf{R@3}$_{256}\uparrow$ & $\boldsymbol{C}_{\textbf{prec}}\uparrow$ & $\boldsymbol{C}_{\textbf{rec}}\uparrow$ & $\textbf{F1}_H\uparrow$ \\
\midrule
w/o feedback & 0.126 & 2.257 & 0.757 & 0.832 & 0.793 & 0.783 \\
\midrule
Self-conditioning & \snd{0.119} & 2.258 & 0.760 & 0.835 & 0.801 & 0.790 \\
Token only & 0.121 & 2.255 & 0.758 & \snd{0.846} & \snd{0.832} & \snd{0.820} \\
Block only & \snd{0.119} & 2.229 & 0.766 & 0.834 & 0.818 & 0.814 \\
Unrouted & 0.123 & \snd{2.189} & \snd{0.769} & 0.834 & 0.824 & 0.816 \\
Both, routed (ours) & \fst{0.114} & \fst{2.154} & \fst{0.779} & \fst{0.855} & \fst{0.846} & \fst{0.834} \\
\bottomrule
\end{tabular}}
\vspace{-3.5mm}
\end{table}

\begin{table}[t]
\setlength{\abovecaptionskip}{2pt}\setlength{\belowcaptionskip}{0pt}
\centering
\vspace{3.5mm}
\caption{\textbf{Hand-object coupling with and without contact feedback.} Pene: mean penetration depth of all joints (mm), as in \cref{tab:main}; $\text{Pene}_H$: the same over the hand joints only; $\text{Contact}_H$: frames with a hand in contact; Floating as in \cref{tab:unseen}.}
\label{tab:consistency}
\vspace{0.5mm}
\resizebox{\columnwidth}{!}{%
\begin{tabular}{l|cccc}
\toprule
\textbf{Variant} & \textbf{Pene}$\downarrow$ & $\textbf{Pene}_H\downarrow$ & $\textbf{Contact}_H\!\to$ & \textbf{Floating}$\downarrow$ \\
\midrule
w/o feedback & \snd{1.045} & \snd{1.603} & \snd{61.2\%} & \snd{9.6\%} \\
With feedback (ours) & \fst{0.941} & \fst{1.259} & \fst{65.0\%} & \fst{7.4\%} \\
\midrule
\rowcolor{rowgray}
Ground truth & 0.623 & 0.954 & 67.0\% & 5.0\% \\
\bottomrule
\end{tabular}}
\end{table}

\paragraph{Hand-object coupling.}
To test if the hands and  object move together, we also measure hand penetration, the hand contact share and floating-object frames (\cref{tab:consistency}). Feedback lowers both penetration measures, brings the hand contact share closer to the ground truth and reduces the floating-object frames.

\begin{figure}[t]
\centering
\vspace{2mm}
\includegraphics[width=\columnwidth]{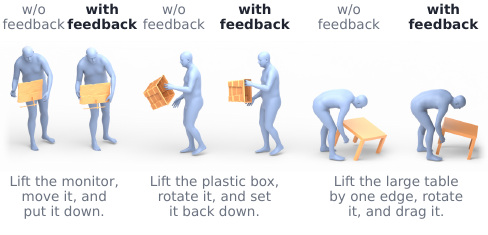}
\vspace{-25pt}
\caption{\textbf{Effect of contact feedback.} Three test prompts sampled from the same noise with the feedback disabled (left of each pair) and enabled (right). Without feedback the hands stay off the monitor, the box and the table edge; with feedback they close onto the object.}
\label{fig:fb}
\vspace{-2mm}
\end{figure}

\begin{table}[t]
\setlength{\abovecaptionskip}{2pt}\setlength{\belowcaptionskip}{0pt}
\centering
\caption{\textbf{Perturbing the feedback at inference.} Feedback disabled, frozen after the first step, or delivered to the wrong part tokens (L$\leftrightarrow$R: hands exchanged; H$\leftrightarrow$F: hands and feet exchanged). $\text{F1}_C$: whole-body contact F1.}
\label{tab:fb_perturb}
\vspace{0.5mm}
\resizebox{\columnwidth}{!}{%
\begin{tabular}{l|cccc|c}
\toprule
& \textbf{Disabled} & \textbf{Frozen} & \textbf{L}$\leftrightarrow$\textbf{R} & \textbf{H}$\leftrightarrow$\textbf{F} & \textbf{Standard (ours)} \\
\midrule
$\textbf{F1}_C\uparrow$ & \snd{0.402} & 0.351 & 0.366 & 0.263 & \fst{0.456} \\
$\textbf{F1}_H\uparrow$ & \snd{0.776} & 0.759 & 0.742 & 0.654 & \fst{0.834} \\
\bottomrule
\end{tabular}}
\vspace{-3.5mm}
\end{table}

\begin{table}[t]
\setlength{\abovecaptionskip}{2pt}\setlength{\belowcaptionskip}{0pt}
\centering
\vspace{3.5mm}
\caption{\textbf{Contact feedback in LIGHT~\cite{light2026} and InterAct~\cite{xu2025interact}.} Each is retrained without and with our feedback.}
\label{tab:light}
\vspace{0.5mm}
\resizebox{\columnwidth}{!}{%
\begin{tabular}{l|c|ccc|c}
\toprule
\textbf{Variant} & \textbf{FID}$\downarrow$ & $\boldsymbol{C}_{\textbf{prec}}\uparrow$ & $\boldsymbol{C}_{\textbf{rec}}\uparrow$ & $\textbf{F1}_H\uparrow$ & \textbf{Floating}$\downarrow$ \\
\midrule
InterAct w/o feedback & \snd{0.322} & \snd{0.758} & \snd{0.723} & \snd{0.713} & \snd{19.5\%} \\
InterAct with feedback & \fst{0.285} & \fst{0.779} & \fst{0.754} & \fst{0.749} & \fst{16.4\%} \\
\midrule
LIGHT w/o feedback & \snd{0.170} & \fst{0.830} & \snd{0.698} & \snd{0.730} & \snd{16.1\%} \\
LIGHT with feedback & \fst{0.158} & \snd{0.829} & \fst{0.741} & \fst{0.758} & \fst{13.2\%} \\
\bottomrule
\end{tabular}}
\vspace{-3.5mm}
\end{table}

\paragraph{Feedback during sampling.}
\Cref{fig:fb} compares samples drawn from the same noise with the feedback disabled and enabled. To test whether the gain comes from the measurement itself, we perturb the feedback of the trained model at inference (\cref{tab:fb_perturb}). Freezing it after the first step drops contact below sampling with the feedback disabled, and delivering each region's measurement to another region's token has the same effect. The model relies on a measurement that is current and that belongs to its own part.

\paragraph{Transfer to other generators.}
To test whether the feedback depends on our architecture, we retrain LIGHT~\cite{light2026} and InterAct~\cite{xu2025interact} with it. Both improve in FID, contact recall and F1 and produce fewer floating-object frames (\cref{tab:light}).

\section{Conclusion}
\label{sec:conclusion}

We presented \ours{}, a text-driven HOI generator in which planning and contact correction form one design. 
CHORD decides the interaction where it can be checked, and checks it where it is decided.
The part tokens give every contact region its own place in the plan, surface feedback revises that place at every flow step, and the completion stage keeps what the plan has settled. \ours{} outperforms prior methods in realism and contact accuracy on InterAct, and its feedback carries over to unseen objects and to existing generators.

\paragraph{Limitations.}
The feedback acts on the planned joints, and the fingers are completed around them without a loop of their own; closing the loop at the finger level is a natural extension. \ours{} also models a single rigid object in a clip of pre-specified, bounded length ($L \le 300$ frames); articulated objects and longer sequences are left for future work.

\end{document}